\documentclass[runningheads]{llncs}
\usepackage{graphicx}
\usepackage{amsmath,amssymb} 
\usepackage{color}

\usepackage{mathtools,xparse}
\usepackage{xspace}
\usepackage{multirow}

\makeatletter
\DeclareRobustCommand\onedot{\futurelet\@let@token\@onedot}
\def\@onedot{\ifx\@let@token.\else.\null\fi\xspace}

\def\eg{\emph{e.g}\onedot} 
\def\ie{\emph{i.e}\onedot}

\def\etal{\emph{et al}\onedot}
\makeatother

\DeclarePairedDelimiter{\norm}{\lVert}{\rVert}
\NewDocumentCommand{\normL}{ s O{} m }{%
  \IfBooleanTF{#1}{\norm*{#3}}{\norm[#2]{#3}}_{L_2(\Omega)}%
}

\begin{document}

\title{Conditional Image-Text Embedding Networks}

\titlerunning{Conditional Image-Text Embedding Networks}

\authorrunning{B.\ A.\ Plummer~\etal}

\author{Bryan A. Plummer$^\dagger$
\and
Paige Kordas$^\dagger$
\and
M. Hadi Kiapour$^\ddagger$
\and 
Shuai Zheng$^\ddagger$
\and 
Robinson Piramuthu$^\ddagger$
\and 
Svetlana Lazebnik$^\dagger$}


\institute{University of Illinois at Urbana-Champaign$^\dagger$\\
	\email{\{bplumme2,pkordas2,slazebni\}@illinois.edu}\\
    Ebay Inc.$^\ddagger$\\
\email{\{mkiapour,shuzheng,rpiramuthu\}@ebay.com}
}

\maketitle              
\begin{abstract}
This paper presents an approach for grounding phrases in images which jointly learns multiple text-conditioned embeddings in a single end-to-end model.  In order to differentiate text phrases into semantically distinct subspaces, we propose a concept weight branch that automatically assigns phrases to embeddings, whereas prior works predefine such assignments.  Our proposed solution simplifies the representation requirements for individual embeddings and allows the underrepresented concepts to take advantage of the shared representations before feeding them into concept-specific layers. Comprehensive experiments verify the effectiveness of our approach across three phrase grounding datasets, Flickr30K Entities, ReferIt Game, and Visual Genome, where we obtain a (resp.) 4\%, 3\%, and 4\% improvement in grounding performance over a strong region-phrase embedding baseline\footnote{Code:  \url{https://github.com/BryanPlummer/cite}}.
\end{abstract}
\section{Introduction}
Phrase grounding attempts to localize a given natural language phrase in an image.  This constituent task has applications to image captioning~\cite{fang2014captions,densecap2015,karpathy2014deep,liuAAAI2017,xu2015show}, image retrieval~\cite{gordoECCV2016,RadenovicECCV2016}, and visual question answering~\cite{AntolICCV2015,tommasiBMVC2016,fukui16emnlp}.  Research on phrase grounding has been spurred by the release of several datasets, some of which primarily contain relatively short phrases~\cite{kazemzadeh-EtAl:2014:EMNLP2014,krishnavisualgenome}, while others contain longer queries, including entire sentences that can provide rich context~\cite{flickrentitiesijcv,mao2016generation}.  The difference in query length compounds the already challenging problem of generalizing to any (including never before seen) natural language input. Despite this, much of the recent attention has focused on learning a single embedding model between image regions and phrases~\cite{fukui16emnlp,mao2016generation,hu2015natural,rohrbach2015,wang2016CVPR,wangTwoBranch2017,yehNIPS2017,Luo_2017_CVPR}.

\begin{figure*}[t]
\centering
\includegraphics[width=\textwidth,trim={0cm 0cm 0cm 0cm},clip]{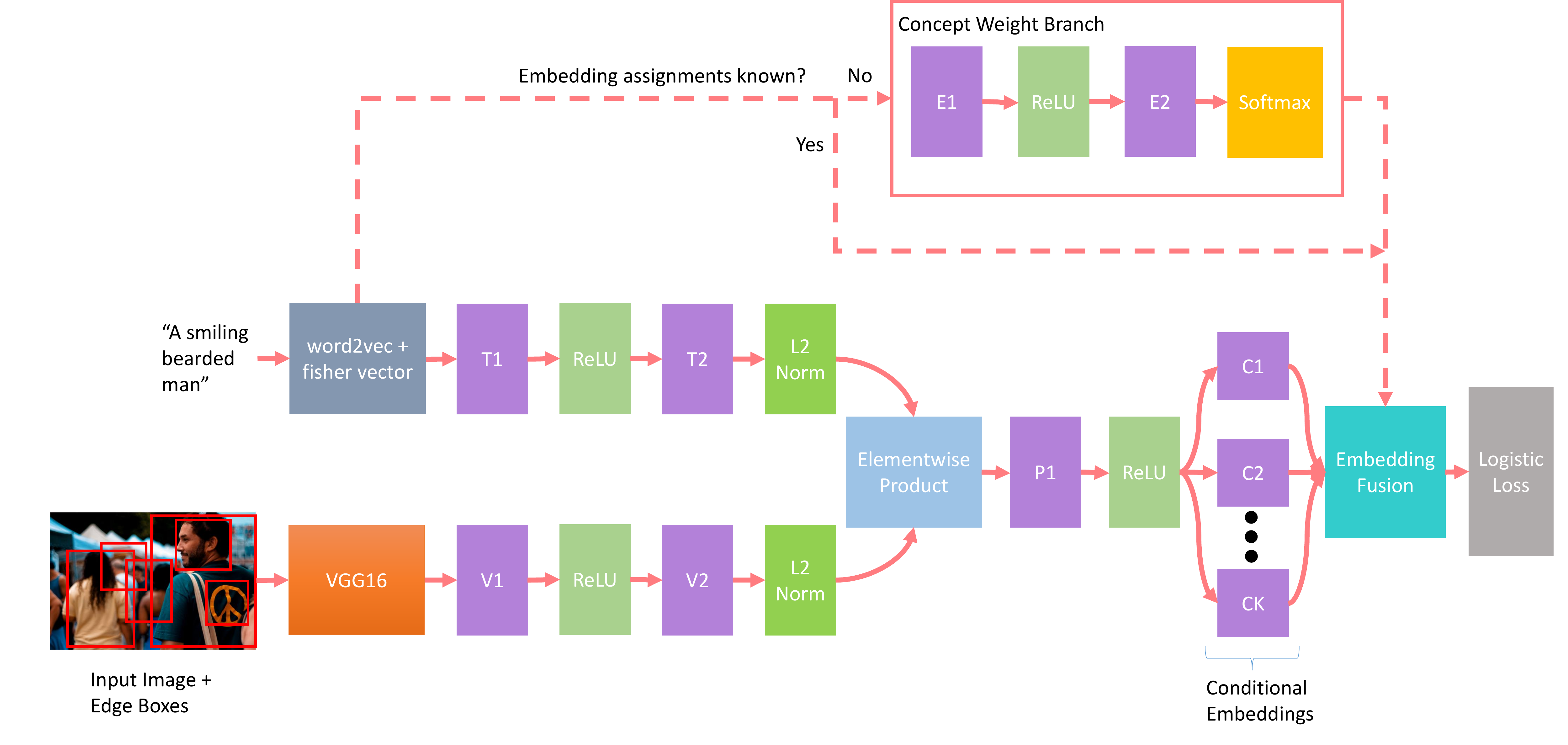}
\caption{Our CITE model separates phrases into different groups and learns conditional embeddings for these groups in a single end-to-end model. Assignments of phrases to embeddings can either be pre-defined (\eg by separating phrases into distinct concepts like \emph{people} or \emph{clothing}), or can be jointly learned with the embeddings using the concept weight branch. Similarly colored blocks refer to layers of the same type, with purple blocks representing fully connected layers.  Best viewed in color}
\label{fig:overview}
\end{figure*}

In this paper, we propose a Conditional Image-Text Embedding (CITE) network that jointly learns different embeddings for subsets of phrases (Figure~\ref{fig:overview}).  This enables our model to train separate embeddings for phrases that share a concept.  
Each conditional embedding can learn a representation specific to a subset of phrases while also taking advantage of weights that are shared across phrases.  This is especially important for smaller groups of phrases that would be prone to overfitting if we were to train separate embeddings for them.  In contrast to similar approaches that manually determine how to group concepts~\cite{Liu_2017_ICCV,plummerPLCLC2017,veitCVPR2017}, we use a concept weight branch, trained jointly with the rest of the network, to do a soft assignment of phrases to learned embeddings automatically.   
The concept weight branch can be thought of producing a unique embedding for each region-phrase pair based on a phrase-specific linear combination of individual conditional embeddings. 
 By training multiple embeddings our model also reduces variance akin to an ensemble of networks, but with far fewer parameters and lower computational cost.  


Our idea of conditional embeddings was directly inspired by the conditional similarity networks of Veit et al.~\cite{veitCVPR2017}, although that work does not deal with cross-modal data and does not attempt to automatically assign different input items to different similarity subspaces. An earlier precursor of the idea of conditional similarity metrics can be found in~\cite{BabenkoICCV2009}.
Our work is also similar in spirit to Zhang~\etal\cite{Zhang_2017_CVPR}, who produced a linear classifier used to discriminate between image regions based on the textual input. 

Our primary focus is on improving methods of associating individual image regions with individual phrases. Orthogonal to this goal, other works have focused on performing global inference for multiple phrases in a sentence and multiple regions in an image.  Wang~\etal~\cite{wang2016matching} modeled the pronoun relationships between phrases and forced each phrase prediction associated with a caption to be assigned to a different region. Chen~\etal~\cite{ChenICMR2017} also took into account the predictions made by other phrases when localizing phrases and incorporated bounding box regression to improve their region proposals.  In their follow-up work~\cite{ChenICCV2017}, they introduced a region proposal network for phrases effectively reproducing the full Faster RCNN detection pipeline~\cite{renNIPS15fasterrcnn}.  Yu~\etal~\cite{yu2016context} took into account the visual similarity of objects in a single image when providing context for their predictions. Plummer~\etal~\cite{plummerPLCLC2017} performed global inference using a wide range of image-language constraints derived from attributes, verbs, prepositions, and pronouns.  Yeh~\etal~\cite{yehNIPS2017} used a word prior in combination with segmentation masks, geometric features, and detection scores to select a region from all possible bounding boxes in an image. Many of these modifications could be used in combination with our approach to further improve performance.

The contributions of our paper are summarized below:

\begin{itemize}
\item By conditioning the embedding used by our model on the input phrase we simplify the representation requirements for each embedding, leading to a more generalizable model.
\item We introduce a concept weight branch which enables our embedding assignments to be learned jointly with the image-text model.
\item We introduce several improvements to the Similarity Network of Wang~\etal~\cite{wangTwoBranch2017} boosting the baseline model's localization performance by 3.5\% over the original paper.
\item We perform extensive experiments over three datasets, Flickr30K Entities~\cite{flickrentitiesijcv}, ReferIt Game~\cite{kazemzadeh-EtAl:2014:EMNLP2014}, and Visual Genome~\cite{krishnavisualgenome}, where we report a (resp.) 4\%, 3\% and 4\% improvement in phrase grounding performance over the baseline.
\end{itemize}

We begin Section~\ref{sec:image-text_embed} by describing the image-text Similarity Network~\cite{wangTwoBranch2017} that we use as our baseline model.  Section~\ref{sec:cond_im_text} describes our text-conditioned embedding model.  Section~\ref{sec:mask_assign} discusses three methods of assigning phrases to the trained embeddings.  Lastly, Section~\ref{sec:experiments} contains detailed experimental results and analysis of our proposed approach.

\section{Our Approach}
\subsection{Image-Text Similarity Network}
\label{sec:image-text_embed}
Given an image and a phrase, our goal is to select the most likely location of the phrase from a set of region proposals.  To accomplish this, we build upon the image-text similarity network introduced in Wang~\etal~\cite{wangTwoBranch2017}.  The image and text branches of this network each have two fully connected layers with batch normalization~\cite{Ioffe:2015:BNA:3045118.3045167} and ReLUs.  The final outputs of these branches are L2 normalized before performing an element-wise product between the image and text representations.  This representation is then fed into a triplet of fully connected layers using batch normalization and ReLUs.  This is analogous to using the CITE model in Figure~\ref{fig:overview} with a single conditional embedding.

The training objective for this network is a logistic regression loss computed over phrases $P$, the image regions $R$, and labels $Y$. The label $y_{ij}$ for the $i$th input phrase and $j$th region is $+1$  where they match and $-1$ otherwise.  Since this is a supervised learning approach, matching pairs of phrases and regions need to be provided in the annotations of each dataset.  After producing some score $x_{ij}$ measuring the affinity between the image region and text features using our network, the loss is given by

\begin{equation}
L_{sim}(P,R,Y) = \sum_{ij}\log(1 + \exp{(-y_{ij}x_{ij})}).
\end{equation}

\noindent In this formulation, it is easy to consider multiple regions for a given phrase as positive examples and to use a variable number of region proposals per image.  This is in contrast to competing methods which score regions with softmax with a cross entropy loss over a set number of proposals per image (\eg~\cite{fukui16emnlp,rohrbach2015,ChenICMR2017}).
\smallskip

\noindent{\bf Sampling phrase-region training pairs.}  Following Wang~\etal~\cite{wangTwoBranch2017}, we consider any regions with at least 0.6 intersection over union (IOU) with the ground truth box for a given phrase as a positive example.  Negative examples are randomly sampled from regions of the same image with less than 0.3 IOU with the ground truth box.  We select twice the number of negative regions as we have positive regions for a phrase.  If too few negative regions occur for an image-phrase pair, then the negative example threshold is raised to 0.4 IOU.
\smallskip

\noindent{\bf Features.} We represent phrases using the HGLMM fisher vector encoding~\cite{klein2014fisher} of word2vec~\cite{mikolov2013efficient} PCA reduced down to 6,000 dimensions. We generate region proposals using Edge Boxes~\cite{ZitnickECCV14}. Similarly to most state-of-the-art methods on our target datasets, we represent image regions using a Fast RCNN network~\cite{girshickICCV15fastrcnn} fine-tuned on the union of PASCAL 2007 and 2012 trainval sets~\cite{pascal-voc-2012}. The only exception is the experiment reported in Table \ref{tab:flickr_overall}(d), where we fine-tune the Fast RCNN parameters (corresponding to the VGG16 box in Figure \ref{fig:overview}) on the Flickr30K Entities dataset.
\smallskip

\noindent{\bf Spatial location.} Following~\cite{rohrbach2015,ChenICMR2017,ChenICCV2017,yu2016context}, we experiment with concatenating bounding box location features to our region representation.  This way our model can learn to bias predictions for phrases based on their location (\eg that \emph{sky} typically occurs in the top part of an image).  For Flickr30K Entities we encode this spatial information as defined in~\cite{ChenICMR2017,ChenICCV2017} for this dataset.  For an image of height $H$ and width $W$ and a box with height $h$ and width $w$ is encoded as $[x_{min}/W, y_{min}/H, x_{max}/W, y_{max}/H, wh/WH]$. For a fair comparison to prior work~\cite{rohrbach2015,ChenICMR2017,ChenICCV2017},  experiments on the ReferIt Game dataset encode the spatial information as an 8-dimensional feature vector $[x_{min}, y_{min}, x_{max}, y_{max}, x_{center},$ $ y_{center}, w, h]$.  For Visual Genome we adopt the same method of encoding spatial location as used for the ReferIt Game dataset.

\subsection{Conditional Image-Text Network}
\label{sec:cond_im_text}
Inspired by Veit~\etal~\cite{veitCVPR2017}, we modify the image-text similarity model of the previous section to learn a set of conditional or concept embedding layers denoted $C_1, \ldots C_K$ in Figure ~\ref{fig:overview}. These are $K$ parallel fully connected layers each with output dimensionality $M$. The outputs of these layers, in the form of a matrix of size $M \times K$, are fed into the embedding fusion layer, together with a $K$-dimensional concept weight vector $U$, which can be produced by several methods, as discussed in Section \ref{sec:mask_assign}. The fusion layer simply performs a matrix-vector product, \ie, $F = CU$. This is followed by another fully connected layer representing the final classifier (\ie, the layer's output dimension is 1).



\subsection{Embedding Assignment}
\label{sec:mask_assign}
This section describes three possible methods for producing the concept weight vector $U$ for combining the conditional embeddings as introduced in Section \ref{sec:cond_im_text}.
\smallskip

\noindent{\bf Coarse categories.}  The Flickr30K Entities dataset comes with hand-constructed dictionaries that group phrases into eight coarse categories: \emph{people, clothing, body parts, animals, vehicles, instruments, scene, other}.  We use these dictionaries to map phrases to binary concept vectors representing their group membership.  This is analogous to the approach of Veit~\etal~\cite{veitCVPR2017}, which defines the concepts based on meta-data labels.  Both the remaining approaches base their assignments on the training data rather than a hand-defined category label.
\smallskip

\noindent{\bf Nearest cluster center.}  A simple method of creating concept weights is to perform K-means clustering on the text features of the queries in the test set.  Each cluster center becomes its own concept to learn.  The concept weights $U$ are then encoded as one-hot cluster membership vectors which we found to work better than alternatives such as similarity of a sample to each cluster center.
\smallskip

\noindent{\bf Concept weight branch.} Creating a predefined set of concepts to learn, either using dictionaries or K-means clustering, produces concepts that don't necessarily have anything to do with the difficulty or ease in localizing the phrases within them.  An alternative is to let the model decide which concepts to learn.  With this in mind, we feed the raw text features into a separate branch of the network consisting of two fully connected layers with batch normalization and a ReLU between them, followed by a softmax layer to ensure the output sums to 1 (denoted as the concept weight branch in Figure~\ref{fig:overview}).  The output of the softmax is then used as the concept weights $U$.  This can be seen as analogous to using soft attention~\cite{xu2015show} on the text features to select concepts for the final representation of a phrase.  We use L1 regularization on the output of the last fully connected layer before being fed into the softmax to promote sparsity in our assignments.  The training objective for our full CITE model then becomes

\begin{equation}
L_{CITE} = L_{sim}(P,R,Y) + \lambda\norm{\phi}_1,
\label{eq:final_loss}
\end{equation}

\noindent where $\phi$ are the inputs to the softmax layer and $\lambda$ is a parameter controlling the importance of the regularization term.  Note that we do not enforce diversity of assignments between different phrases, so it is possible that all phrases attend to a single embedding.  However, we do not see this actually occur in practice.  We also tried to use entropy minimization rather then L1 regularization for our concept weight branch as well as hard attention instead of soft attention, but found all worked similarly in our experiments.

\section{Experiments}
\label{sec:experiments}

\subsection{Datasets and Protocols}

We evaluate the performance of our phrase-region grounding model on three datasets: Flickr30K Entities~\cite{flickrentitiesijcv}, ReferIt Game~\cite{kazemzadeh-EtAl:2014:EMNLP2014}, and Visual Genome~\cite{krishnavisualgenome}. The metric we report is the proportion of correctly localized phrases in the test set. Consistent with prior work, a 0.5 IOU between the best-predicted box for a phrase and its ground truth is required for a phrase to be considered successfully localized.    Similarly to~\cite{wangTwoBranch2017,plummerPLCLC2017,ChenICCV2017}, for phrases associated with multiple bounding boxes, the phrase is represented as the union of its boxes.
\smallskip

\noindent{\bf Training procedure.} We begin training our models with Adam~\cite{adam}.  After every epoch, we evaluate our model on the validation set. After it hasn't improved performance for 5 epochs, we fine-tune our model with stochastic gradient descent at 1/10th the learning rate and the same stopping criteria.  We report test set performance for the model that performed best on the validation set.
\smallskip

\noindent{\bf Comparative evaluation.} In addition to comparing to previously published numbers of state-of-the-art approaches on each dataset, we systematically evaluate the following baselines and variants of our model:

\begin{itemize}
\item {\bf Similarity Network.}  Our first baseline is given by our own implementation of the model from Wang~\etal~\cite{wangTwoBranch2017}, trained using the procedure described above. Phrases are pre-processed using stop word removal rather than part-of-speech filtering as done in the original paper.  This change, together with a more careful tuning of the training settings, leads to a 2.5\% improvement in performance over the reported results in~\cite{wangTwoBranch2017}.  The model is further enhanced by using the spatial location features (Section \ref{sec:image-text_embed}), resulting in a total improvement of 3.5\%.
\item {\bf Individual Coarse Category Similarity Networks.}  We train multiple Similarity Networks on different subsets of the data created according to the coarse category assignments as described in Section~\ref{sec:mask_assign}.
\item {\bf Individual K-means Similarity Networks.}  We train multiple Similarity Networks on different subsets of the data created according to the nearest cluster center assignments as described in Section~\ref{sec:mask_assign}.
\item {\bf CITE, Coarse Categories.} No concept weight branch. Phrases are assigned according to their coarse category.
\item {\bf CITE, Random.} No concept weight branch. Phrases are randomly assigned to an embedding.  At test time, phrases seen during training keep their assignments, while new phrases are randomly assigned.
\item {\bf CITE, K-means.}  No concept weight branch. Phrases are matched to embeddings using nearest cluster center assignments.
\item {\bf CITE, Learned.}  Our full model with the concept weight branch used to automatically produce concept weights as described in Section~\ref{sec:mask_assign}.
\end{itemize}

\subsection{Flickr30K Entities}
\label{sec:flickr}

\begin{table}[t]
  \centering
  \caption{Phrase localization performance on the Flickr30k Entities test set. (a) State-of-the-art results when predicting a single phrase at a time taken from published works. 
(b,c) Our baselines and variants using PASCAL-tuned features. (d) Results using Flickr30k-tuned features}
\label{tab:flickr_overall}
    \begin{tabular}{|ll|c|}
      \hline
      & Method & Accuracy\\
      \hline
      \hline
      (a) & {\bf Single Phrase Methods (PASCAL-tuned Features)\footnotemark} & \\
      & NonlinearSP~\cite{wang2016CVPR} &  43.89 \\
      & GroundeR~\cite{rohrbach2015} &  47.81 \\
      & MCB~\cite{fukui16emnlp} &  48.69 \\
      & RtP~\cite{flickrentitiesijcv} &  50.89 \\
      & Similarity Network
      ~\cite{wangTwoBranch2017} 
      &  51.05 \\
      & IGOP~\cite{yehNIPS2017} & 53.97\\
      & SPC~\cite{plummerPLCLC2017} & 55.49\\
      & MCB + Reg + Spatial~\cite{ChenICMR2017} & 51.01\\
      & MNN + Reg + Spatial~\cite{ChenICMR2017} & 55.99\\
      \hline
      (b) & {\bf Our Implementation} &\\
      & Similarity Network & 53.45 \\
      & Similarity Network + Spatial & 54.52 \\
      \hline
      (c) & {\bf Conditional Models + Spatial} &\\
      & Individual Coarse Category Similarity Networks, $K=8$ & 55.32\\
      & Individual K-means Similarity Networks, $K=8$ & 54.95\\
      & CITE, Coarse Categories, $K=8$ & 55.42\\
      & CITE, Random, $K=16$ & 57.58\\
      & CITE, K-means, $K=16$ & 57.89\\
      & CITE, Learned, $K=4$ & 58.69\\
      & CITE, Learned, $K=4$, 500 Edge Boxes & 59.27\\
      \hline
      (d) & {\bf Flickr30K-tuned Features + Spatial} &\\
      & PGN + QRN~\cite{ChenICCV2017} & 60.21\\
      & CITE, Learned, $K=4$, 500 Edge Boxes & {\bf 61.89}\\
      \hline
    \end{tabular}
\end{table}

We use the same splits as Plummer~\etal~\cite{flickrentitiesijcv}, which separates the images into 29,783 for training, 1,000 for testing, and 1,000 for validation.  Models are trained with a batch size of 200 (128 if necessary to fit into GPU memory) and learning rate of 5e-5. We set $\lambda =$ 5e-5 in Eq.~(\ref{eq:final_loss}).  We use the top 200 Edge Box proposals per image and embedding dimension $M=256$ unless stated otherwise.
\smallskip

\footnotetext{Performance on this task can be further improved by taking into account the predictions made for other phrases in the same sentence~\cite{plummerPLCLC2017,wang2016matching,ChenICMR2017,ChenICCV2017}, with the best result using Pascal-tuned features of 57.53\% achieved by Chen~\etal~\cite{ChenICMR2017} and 65.14\% using Flickr30K-tuned features~\cite{ChenICCV2017}.}

\noindent{\bf Grounding Results.} Table~\ref{tab:flickr_overall} compares overall localization accuracies for a number of methods. The numbers for our Similarity Network baseline are reported in Table~\ref{tab:flickr_overall}(b), and as stated above, they are better than the published numbers from~\cite{wangTwoBranch2017}. Table~\ref{tab:flickr_overall}(c) reports results for variants of conditional embedding models. From the first two lines, we can see that learning embeddings from subsets of the data without any shared weights leads to only a small improvement ($\leq$ 1\%) over the Similarity Network baseline.  The third line of Table~\ref{tab:flickr_overall}(c)  reports that separating phrases by manually defined high-level concepts only leads to a 1\% improvement even when weights are shared across embeddings.  This is likely due, in part, to the significant imbalance between different coarse categories, as a uniform random assignment shown in the fourth line of Table~\ref{tab:flickr_overall}(c) lead to a 3\% improvement.  The fifth line of Table~\ref{tab:flickr_overall}(c) demonstrates that grouping phrases based on their text features better reflects the needs of the data, resulting in just over 3\% improvement over the baseline, only slightly better than random assignments.  An additional improvement is reported in the eighth line of Table~\ref{tab:flickr_overall}(c) by incorporating our concept weight branch, enabling our model to both determine what concepts are important to learn and how to assign phrases to them.  We see in the last line of Table~\ref{tab:flickr_overall}(c)  that going from 200 to 500 bounding box proposals provides a small boost in localization accuracy. This results in our best performance using PASCAL-tuned features which is 3\% better than the prior work reported in Table~\ref{tab:flickr_overall}(a) and 4.5\% better than the Similarity Network.  We also note that the time to test an image-phrase pair is almost unaffected using our approach (the CITE, Learned, K=4 model performs inference on 200 Edge Boxes at 0.182 seconds per pair using a NVIDIA Titan X GPU with our implementation) compared with the baseline Similarity Network (0.171 seconds per pair). Finally, Table~\ref{tab:flickr_overall}(d) gives results for models whose visual features were fine-tuned for localization on the Flickr30K Entities dataset.  Our model still obtains a 1.5\% improvement over the approach of Chen~\etal~\cite{ChenICCV2017}, which used bounding box regression as well as a region proposal network. In principle, we could also incorporate these techniques to further improve the model.

Table~\ref{tab:type_results} breaks down localization accuracy by coarse category.  Of particular note are our results on the challenging {\em body part} category, which are typically small and represent only 3.5\% of the phrases in the test set, improving over the next best model as well as the Similarity Network trained on just body part phrases by 10\% when using Flickr30K-tuned features.  We also see a substantial improvement in the {\em vehicles} and {\em other} categories, seeing a 5-9\% improvement over the previous state-of-the-art.  The only category where we perform worse are phrases referring to scenes, which commonly cover the majority (or entire) image. Here, incorporating a bias towards selecting larger proposals, as in~\cite{flickrentitiesijcv,plummerPLCLC2017}, can lead to significant improvements.
\smallskip

\begin{table*}[t]
\centering
  \caption{Comparison of phrase grounding performance over coarse categories on the Flickr30K Entities dataset.  Our models were tested with 500 Edge Box proposals}
    \label{tab:type_results}
  \setlength{\tabcolsep}{1pt}
    \begin{tabular}{|l|c|c|c|c|c|c|c|c|}
      \hline
     & \multirow{2}{*}{People} & Cloth- & Body & Anim- & Vehi- & Instru- & \multirow{2}{*}{Scene} & \multirow{2}{*}{Other} \\
     & & ing & Parts & als & cles & ments &  &  \\
      \hline
      \hline
      \multicolumn{9}{|l|}{PASCAL-tuned Features} \\
        \hline
     GroundeR~\cite{rohrbach2015} & 61.00 & 38.12 & 10.33 & 62.55 & 68.75 & 36.42 & 58.18 & 29.08 \\
     RtP~\cite{flickrentitiesijcv} & 64.73 & 46.88 & 17.21 & 65.83 & 68.75 & 37.65 & 51.39 & 31.77 \\
      IGOP~\cite{yehNIPS2017} & 68.71 & {\bf 56.83} & 19.50 & 70.07 & 73.75 & 39.50 & 60.38 & 32.45\\
      MCB + Reg + Spatial~\cite{ChenICMR2017} & 62.75 & 43.67 & 14.91 & 65.44 & 65.25 & 24.74 & {\bf 64.10} & 34.62\\
      MNN + Reg + Spatial~\cite{ChenICMR2017} & 67.38 & 47.57 & 20.11 & 73.75 & 72.44 & 29.34 & 63.68 & 37.88\\
     CITE, Learned, $K=4$ + Spatial & {\bf 73.20} & 52.34 & {\bf 30.59} & {\bf 76.25} & {\bf 75.75} & {\bf 48.15} & 55.64 & {\bf 42.83}\\
      \hline
  \multicolumn{9}{|l|}{Flickr30K-tuned Features} \\
  \hline
  PGN + QRN + Spatial~\cite{ChenICCV2017} & 75.05 & 55.90 & 20.27 & 73.36 & 68.95 & 45.68 & {\bf 65.27} & 38.80\\
   CITE, Learned, $K=4$ + Spatial& {\bf 75.95} & {\bf 58.50} & {\bf 30.78} & {\bf 77.03} & {\bf 79.25} & {\bf 48.15} & 58.78 & {\bf 43.24}\\
      \hline
    \end{tabular}
\end{table*}

\begin{figure}[t]
\centering
\begin{tabular}{lr}
\includegraphics[width=0.45\textwidth]{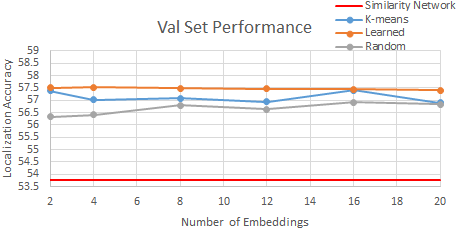}
\includegraphics[width=0.45\textwidth]{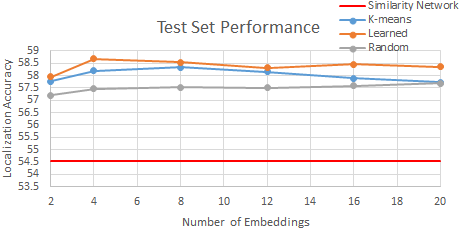}
\end{tabular}
\caption{Effect of the number of learned embeddings ($K$) on Flickr30K Entities localization accuracy using PASCAL-tuned features}
\label{fig:choice_of_K}
\end{figure}

\noindent{\bf Parameter Selection.} In addition to reporting the localization performance, we also provide some insight into the effect of different parameter choices and what information our model is capturing.  In Figure~\ref{fig:choice_of_K} we show how the number $K$ of learned embeddings affects performance.  Using our concept weight branch consistently outperforms K-means cluster assignments.  Table~\ref{tab:embedding_size} shows how the embedding dimensionality $M$ affects performance.  Here we see that reducing the output dimension from 256 to 64 (\ie, by 1/4th) leads to a minor (1\%) decrease in performance.  This result is particularly noteworthy as the CITE network with $K=4, M=64$ has 4 million parameters compared the 14 million the baseline Similarity Network has with $M=256$ while still maintaining a 3\% improvement in performance.  We also experimented with different ways of altering the Similarity Network to have the same number of parameters to ours at similar points (\eg increasing the last fully connected layer to be $K$ times larger or adding $K$ additional layers), but found they performed comparably to the baseline Similarity Network (\ie their performance was about 4\% worse than our approach).  In addition to experiments on how many layers to use and  the size of each layer, we also explored the effect the number of Edge Boxes has on performance in Table~\ref{tab:num_proposals}.  In contrast to some prior work which performed best using 200 candidates (\eg~\cite{flickrentitiesijcv,plummerPLCLC2017}), our model's increased discriminate power enables us to still be able to obtain a benefit from using up to 500 proposals.
\smallskip

\begin{table}[t]
\centering
\caption{Localization accuracy with different embedding sizes using the CITE, Learned, $K = 4$ model on Flickr30K Entities with PASCAL-tuned features.  Embedding size refers to $M$, the output dimensionality of layers P1 and the conditional embeddings in Figure~\ref{fig:overview}.  The remaining fully connected layers' output dimensions (excluding those that are part of the VGG16 network) are four times the embedding size}
\label{tab:embedding_size}
\begin{tabular}{|l|c|c|c|c|}
\hline
Embedding Size ($M$) & 64 & 128 & 256 & 512\\
\hline
Validation Set Accuracy & 56.32 & 57.51 & {\bf 57.53} & 57.42 \\
\hline
Test Set Accuracy & 57.77 & 58.48 & {\bf 58.69} & 58.64\\
\hline
\end{tabular}
\end{table}

\begin{table}[t]
\centering
\caption{Localization accuracy with different numbers of proposals using the CITE, Learned, $K = 4$ model on Flickr30K Entities with PASCAL-tuned features}
\label{tab:num_proposals}
\begin{tabular}{|l|c|c|c|c|}
\hline
\#Edge Box Proposals & 100 & 200 & 500 & 1000\\
\hline
Validation Set Accuracy & 49.61 & 57.53 & 58.48  & 57.87\\
\hline
Test Set Accuracy & 51.32 & 58.69 & 59.27 & 58.63\\
\hline
\end{tabular}
\end{table}

\begin{figure*}[t]
\centering
\begin{tabular}{ccc}
\includegraphics[width=0.45\textwidth]{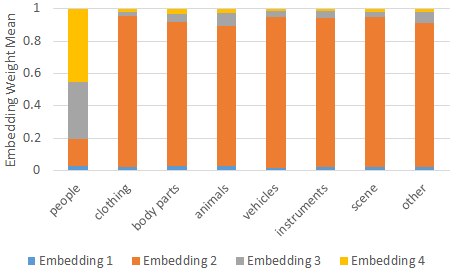} & \hspace{3cm}&\includegraphics[width=0.45\textwidth]{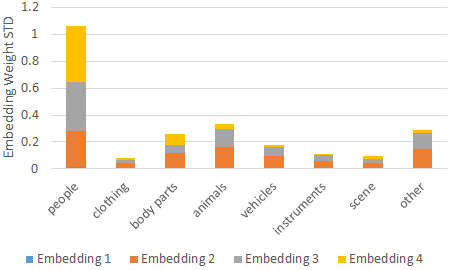}
\end{tabular}
\caption{The mean weight for each embedding (left) along with the standard deviation of those weights (right) broken down by coarse category for the Flickr30K Entities dataset using Flickr30K-tuned features}
\label{fig:weight_graph}
\end{figure*}

\noindent{\bf Concept Weight Branch Examination.} To analyze what our model is learning, Figure~\ref{fig:weight_graph} shows the means and standard deviations of the weights over the different embeddings broken down by coarse categories. Interestingly, {\em people} end up being split between two embeddings. We find that people phrases tend to be split by plural vs.\ singular.  Table~\ref{tab:weight_phrase_list} gives a closer look at the conditional embeddings by listing the ten phrases with the highest weight for each embedding.  While most phrases give the first embedding little weight, it appears to provide the most benefit for finding very specific references to people rather than generic terms (\eg \emph{little curly hair girl} instead of \emph{girl} itself).  These patterns generally hold through multiple runs of the model, indicating they are important concepts to learn for the task.
\smallskip

\begin{table*}[t]
\centering
\caption{The ten phrases with the highest weight per embedding on the Flickr30K Entities dataset using Flickr30K-tuned features}
\label{tab:weight_phrase_list}
\begin{tabular}{|c||p{9.25cm}|}
\hline
Embedding 1 & soldiers (0.08), male nun (0.07), rather angry looking woman (0.07), skinny dark complected boy (0.07), little curly hair girl (0.07), middle eastern woman (0.07), first man's leg (0.07), statue athletic man (0.07), referee (0.07), woman drink wine (0.07)\\
\hline
Embedding 2 & red scooter (0.97), blue clothes (0.97), yellow bike (0.97), red bike (0.97), red buckets (0.97), yellow backpack (0.97), street window shops (0.97), red blue buckets (0.97), red backpack (0.97), purple red backpack (0.97)\\
\hline
Embedding 3 & two people (0.94), two men (0.93), two young kids (0.93), two kids (0.93), two white-haired women (0.93), two women (0.93), group three boys (0.93), two young people (0.93), three people (0.92), crowd people (0.92)\\
\hline
Embedding 4 & blond-haired woman (0.91), dark-skinned woman (0.91), gray-haired man (0.91), one-armed man (0.91), dark-haired man (0.91), red-haired man (0.91), boy young man (0.91), man (0.91), well-dressed man (0.91), dark-skinned man (0.91)\\
\hline
\end{tabular}
\end{table*}

\begin{figure*}[t]
\centering
\includegraphics[width=0.8\textwidth,trim=0cm 0.6cm 0cm 0cm,clip]{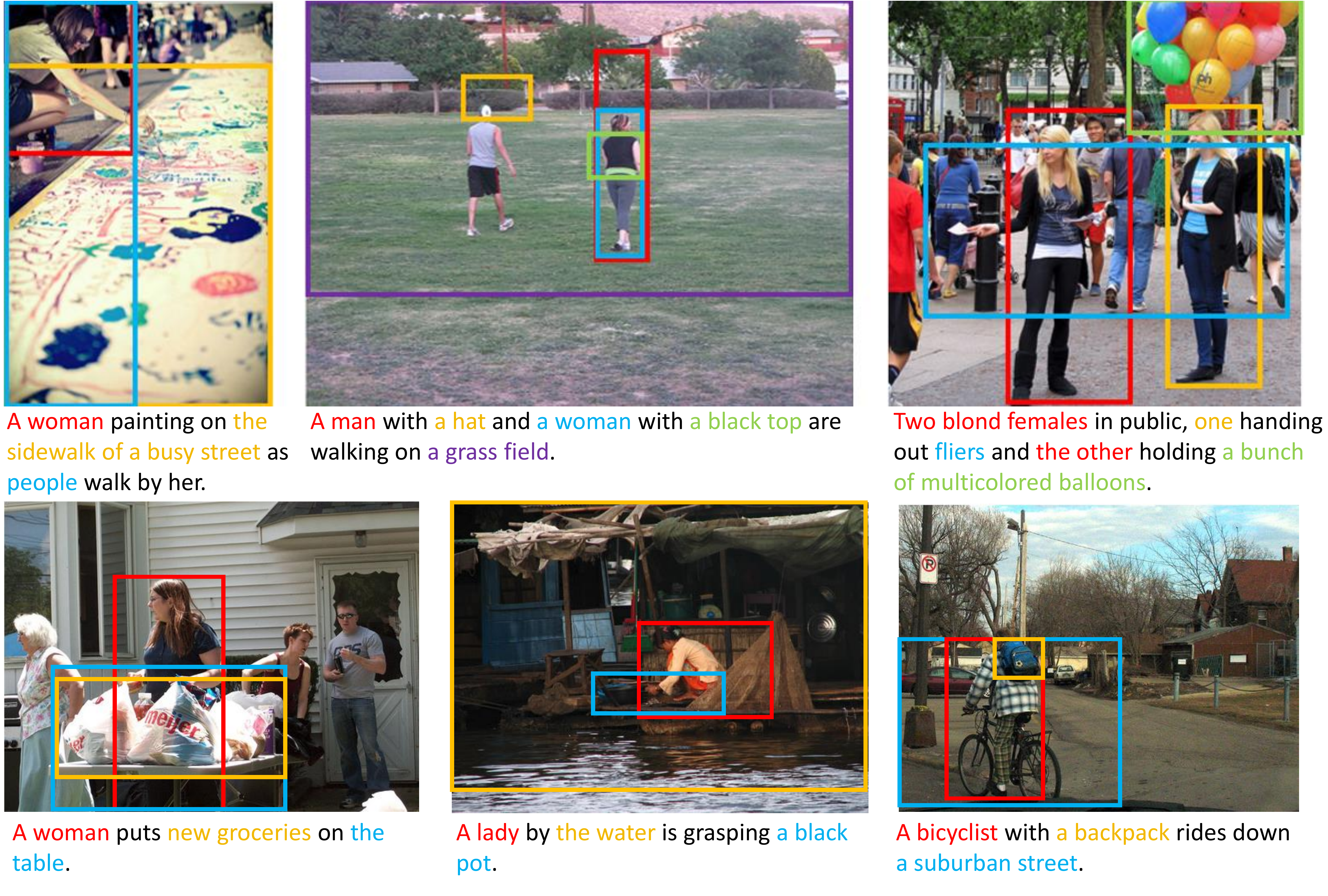}
\caption{Examples demonstrating some common failure cases on the Flickr30K Entities dataset.  See Section~\ref{sec:flickr} for discussion}
\label{fig:flickr_failures}
\end{figure*}

\noindent{\bf Qualitative Results.} Figure~\ref{fig:flickr_failures} gives a look into areas where our model could be improved.  Of the phrases that occur at least 100 times in the test set, the lowest performing phrases are \emph{street} and \emph{people} at (resp.) 60\% and 64\% accuracy.  The highest performing of these common phrases is \emph{man} at 81\% accuracy, which also happens to be the most common phrase with 1065 instances in the test set.  In the top-left example of Figure~\ref{fig:flickr_failures}, the word {\em people}, which is not correctly localized, refers to partially visible background pedestrians.  Analyzing the saliency of a phrase in the context of the whole caption may lead to treating these phrases differently. Global inference constraints, for example, a requirement that predictions for {\em a man} and {\em a woman} must be different, would be useful for the top-center example.  Performing pronoun resolution, as attempted in~\cite{plummerPLCLC2017}, would help in the top-right example.  In the test set, the pronoun \emph{one} is correctly localized around 36\% of the time, whereas \emph{the blond woman} is correctly localized 81\% of the time.  Having an understanding of relationships between entities may help in cases such as the bottom-left example of Figure~\ref{fig:flickr_failures}, where the extent of the table could be refined by knowing that the groceries are ``on'' it.  Our model also performs relatively poorly on phrases referring to classic ``stuff'' categories, as shown in the bottom-center and bottom-right examples.  The {\em water} and {\em street} phrases in these examples are only partly localized.  Using pixel-level predictions may help to recover the full extent of these types of phrases since the parts of the images they refer to are relatively homogeneous.

\subsection{ReferIt Game}
We use the same splits as Hu~\etal~\cite{hu2015natural}, which consist of 10,000  images combined for training and validation with the remaining 10,000 images for testing.  Models are trained with a batch size of 128, learning rate of 5e-4, and $\lambda =$ 5e-4 in Eq.~(\ref{eq:final_loss}).  We generate 500 Edge Box proposals per image.
\smallskip

\noindent{\bf Results.} Table~\ref{tab:referit} reports the localization accuracy across the ReferIt Game test set. The first line of Table~\ref{tab:referit}(b) shows that our model using the nearest cluster center assignments results in a 2.5\% improvement over the baseline Similarity Network.  Using our concept weight branch in order to learn assignments yields an additional small improvement.

We note that we do not outperform the approach of Yeh~\etal~\cite{yehNIPS2017} on this dataset.  This can likely be attributed to the failures of Edge Boxes to produce adequate proposals on the ReferIt Game dataset.  Oracle performance using the top 500 proposals is 93\% on Flickr30K Entities, while it is only 86\% on this dataset.  As a result, the specialized bounding box methods used by Yeh~\etal as well as Chen~\etal~\cite{ChenICMR2017} may play a larger role here.  Our model would also likely benefit from these improved bounding boxes.

\begin{table}[t]
\centering
  \caption{Localization performance on the ReferIt Game test set. (a) Published results and our Similarity Network baseline. (b) Our best-performing conditional models}
\label{tab:referit}
    \begin{tabular}{|ll|c|}
      \hline
      & Method & Accuracy\\
      \hline
      \hline
      (a) & {\bf State-of-the-art} & \\
      & SCRC~\cite{hu2015natural} & 17.93\\
      & GroundeR + Spatial~\cite{rohrbach2015} & 26.93\\
      & MCB + Reg + Spatial~\cite{ChenICMR2017} & 26.54\\
      & CGRE~\cite{Luo_2017_CVPR} & 31.85\\
      & MNN + Reg + Spatial~\cite{ChenICMR2017} & 32.21\\
      & IGOP~\cite{yehNIPS2017} & 34.70\\
      & Similarity Network + Spatial & 31.26\\
      \hline
      (b) & {\bf Conditional Models + Spatial} &\\
      & CITE, K-Means, $K=2$ & 34.01\\
      & CITE, Learned, $K=12$ & 34.13\\
      \hline
    \end{tabular}
\end{table}

\begin{figure}[t]
\centering
\begin{tabular}{lr}
\includegraphics[width=0.45\textwidth]{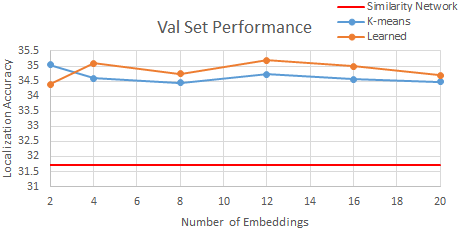}
\includegraphics[width=0.45\textwidth]{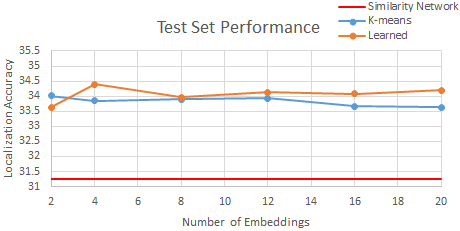}
\end{tabular}
\caption{Effect of the number $K$ of embeddings on localization accuracy on the ReferIt Game dataset}
\label{fig:choice_of_K_referit}
\end{figure}

As with the Flickr30K Entities dataset, we show the effect of the number $K$ of embeddings on localization performance in Figure~\ref{fig:choice_of_K_referit}.  While the concept weight branch provides a small performance improvement across many different choices of K, when $K=2$ the clustering assignments actually perform a little better.  However, this behavior is atypical in our experiments across all three datasets, and may simply be due to the small size of the ReferIt Game training data, as it has far fewer ground truth phrase-region pairs to train our models with.

\subsection{Visual Genome}
We use the same splits as Zhang~\etal~\cite{Zhang_2017_CVPR}, consisting of 77,398 images for training and 5,000 each for testing and validation.  Models are trained with a learning rate of 5e-5, and $\lambda =$ 5e-4 in Eq.~(\ref{eq:final_loss}).  We generate 500 Edge Box proposals per image, and use a batch size of 128.
\smallskip

\noindent{\bf Results.}  Table~\ref{tab:visual_genome} reports the localization accuracy across the Visual Genome dataset. Table~\ref{tab:visual_genome}(a) lists published numbers from several recent methods. The current state of the art performance belongs to Zhang~\etal~\cite{Zhang_2017_CVPR}, who fine-tuned visual features on this dataset and created a cleaner set during training by pruning ambiguous phrases. We did not perform either fine-tuning or phrase pruning, so the most comparable reference number for our methods is their 17.5\% accuracy without these steps. 

The baseline accuracies for our Similarity Network with and without spatial features are given in the last two lines of Table~\ref{tab:visual_genome}(a). We can see that including the spatial features gives only a small improvement.  This is likely due to the denser annotations in this dataset as compared to Flickr30K Entities.  For example, a phrase like \emph{a man} in Flickr30K Entities would typically refer to a relatively large region towards the center since background instances are commonly not mentioned in an image-level caption.  However, entities in Visual Genome include both foreground and background instances.  

In the first line of Table~\ref{tab:visual_genome}(b), we see our K-means model is 3.5\% better than the Similarity Network baseline, and over 6\% better than the 17.5\% accuracy of~\cite{Zhang_2017_CVPR}. According to the second line of Table~\ref{tab:visual_genome}(b), using the concept weight branch obtains a further improvement.  In fact, our full model with pre-trained PASCAL features has better performance than~\cite{Zhang_2017_CVPR} with fine-tuned features. 

\begin{table}[t]
\centering
  \caption{Phrase localization performance on Visual Genome. (a) Published results and our Similarity Network baselines. APP refers to ambiguous phrase pruning (see~\cite{Zhang_2017_CVPR} for details). (b) Our best-performing conditional models}
\label{tab:visual_genome}
    \begin{tabular}{|ll|c|}
      \hline
      & Method & Accuracy\\
      \hline
      \hline
      (a) & {\bf State-of-the-art} & \\
      & Densecap~\cite{Johnson2015CVPR} & 10.1\\
      & SCRC~\cite{hu2015natural} & 11.0\\
      & DBNet~\cite{Zhang_2017_CVPR}  & 17.5\\
      & DBNet (with APP)~\cite{Zhang_2017_CVPR} & 21.2\\
      & DBNet (with APP, V. Genome-tuned Features)~\cite{Zhang_2017_CVPR} & 23.7\\
      & Similarity Network & 19.76\\
      & Similarity Network + Spatial & 20.08\\
      \hline
      (b) & {\bf Conditional Models + Spatial} &\\
      & CITE, K-Means, $K=12$ & 23.67\\
      & CITE, Learned, $K=12$ & 24.43\\
      \hline
    \end{tabular}
\end{table}

\begin{figure}[t]
\centering
\begin{tabular}{lr}
\includegraphics[width=0.45\textwidth]{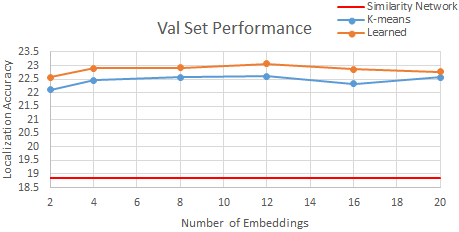}
\includegraphics[width=0.45\textwidth]{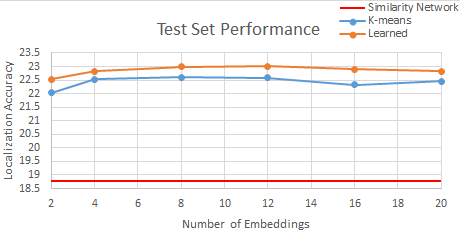}
\end{tabular}
\caption{Effect of the number of learned embeddings on performance on the Visual Genome with models trained on 1/3 of the available training data}
\label{fig:choice_of_K_genome}
\end{figure}

As with the other two datasets, Figure~\ref{fig:choice_of_K_genome} reports performance as a function of the number of learned embeddings.  Echoing most of the earlier results, we see a consistent improvement for the learned embeddings over the K-means ones.  The large size of this dataset ($>$ 250,000 instances in the test set) helps to reinforce the significance of our results.

\section{Conclusion}

This paper introduced a method of learning a set of conditional embeddings and phrase-to-embedding assignments in a single end-to-end network.  The effectiveness of our approach was demonstrated on three popular and challenging phrase-to-region grounding datasets.  In future work, our model could be further improved by including a term to enforce that distinct concepts are being learned by each embedding. 

Our experiments focused on localizing individual phrases to a fixed set of category-independent region proposals. As such, our absolute accuracies could be further improved by incorporating a number of orthogonal techniques used in competing work. By jointly predicting multiple phrases in an image our model could take advantage of relationships between multiple entities (\eg~\cite{plummerPLCLC2017,wang2016matching,ChenICMR2017,ChenICCV2017}).  Including bounding box regression and a region proposal network as done in~\cite{ChenICMR2017,ChenICCV2017} would also likely lead to a better model.  In fact, tying the regression parameters to a specific concept embedding may further improve performance since it would simplify our prediction task as a result of needing to learn parameters for just the phrases assigned to that embedding.
\smallskip

\noindent\textbf{Acknowledgements:} This material is based upon work supported in part by the National Science Foundation under Grants No. 1563727 and 1718221, Amazon Research Award, AWS Machine Learning Research Award, and Google Research Award. 

%
%
%
\bibliographystyle{splncs04}
\bibliography{egbib}
\end{document}